\pdfoutput=1
\documentclass[runningheads]{llncs}
\usepackage{graphicx}
\usepackage{amsmath,amssymb} % define this before the line numbering.
\usepackage{color}
\usepackage{times}
\usepackage{epsfig}
\usepackage{graphicx}
\usepackage{amsmath}
\usepackage{amssymb}

\usepackage{color}
\usepackage{cite}
\usepackage{multirow}
\usepackage{booktabs}
\usepackage{subfigure}
\usepackage{caption}
\usepackage[width=122mm,left=12mm,paperwidth=146mm,height=193mm,top=12mm,paperheight=217mm]{geometry}
\begin{document}
% \linenumbers
\pagestyle{empty}

\mainmatter
\title{A Richly Annotated Dataset for Pedestrian \\ Attribute Recognition} % Replace with your title

\author{Dangwei Li$^{1}$, Zhang Zhang$^{1}$, Xiaotang Chen$^{1}$, Haibin Ling$^{2}$, Kaiqi Huang$^{1}$
}
\institute{$^{1}$CRIPAC$\;\&\;$NLPR, Institute of Automation, Chinese Academy of Sciences\\
%$^{2}$CAS Center for Excellence in Brain Science and Intelligence Technology\\
$^{2}$Computer and Information Sciences Department, Temple University, Philadephia, PA USA}

%\titlerunning{RAP dataset}
%
%\authorrunning{Dangwei Li. etc.}

\maketitle

\begin{abstract}
In this paper, we aim to improve the dataset foundation for pedestrian attribute recognition in real surveillance scenarios. Recognition of human attributes, such as gender, and clothes types, has great prospects in real applications. However, the development of suitable benchmark datasets for attribute recognition remains lagged behind. Existing human attribute datasets are collected from various sources or an integration of pedestrian re-identification datasets. Such heterogeneous collection poses a big challenge on developing high quality fine-grained attribute recognition algorithms. Furthermore, human attribute recognition are generally severely affected by environmental or contextual factors, such as viewpoints, occlusions and body parts, while existing attribute datasets barely care about them. To tackle these problems, we build a Richly Annotated Pedestrian (RAP) dataset\footnote{The website of RAP is \url{http://rap.idealtest.org/}} from real multi-camera surveillance scenarios with long term collection, where data samples are annotated with not only fine-grained human attributes but also environmental and contextual factors. RAP has in total 41,585 pedestrian samples, each of which is annotated with 72 attributes as well as viewpoints, occlusions, body parts information. To our knowledge, the RAP dataset is the largest pedestrian attribute dataset, which is expected to greatly promote the study of large-scale attribute recognition systems. Furthermore, we empirically analyze the effects of different environmental and contextual factors on pedestrian attribute recognition. Experimental results demonstrate that viewpoints, occlusions and body parts information could assist attribute recognition a lot in real applications.
\end{abstract}

\section{Introduction}

% storage, retrieval efficiency, language description.
Recently, recognition of human attributes, such as gender, glasses and clothes types, has drawn a large amount of research attention recently due to its great potential in real surveillance system.
For example, attribute has been used to assists human detection\cite{tian2014pedestrian}, person re-identification \cite{layne2012person,zhu2015multi}, face recognition \cite{kumar2011describable,liu2014deep,luo2013deep}, and has been shown to greatly improve other vision related tasks\cite{ouyang2015deepattributes,shao2015deeply}.
As a middle-level representation, attribute may bridge the gap between low-level features and human description \cite{vaquero2009attribute,feris2014attribute}. In addition, attribute may play the critical role for people search in practical applications for anti-terrorism \cite{feris2014attribute}, such as the retrieval of the two suspects in Boston marathon bombing event.

\begin{figure}
\begin{center}
\includegraphics[width=0.85\linewidth]{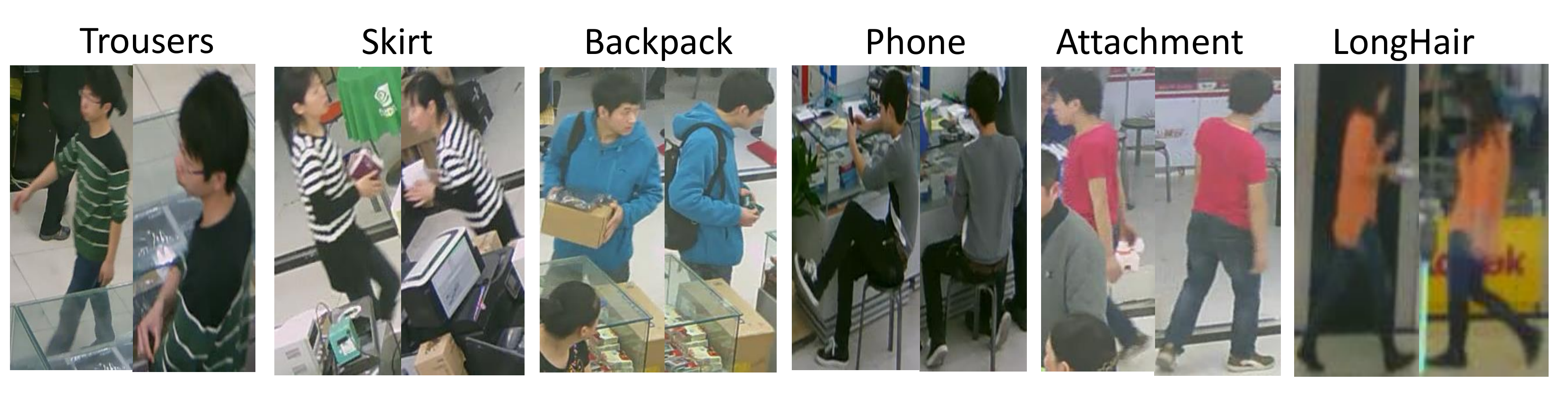}
\vspace{-1.8mm}\caption{Examples of different attributes in the RAP dataset. In real scene, attribute (from left to right) will change or be hard to determine due to camera viewpoint, body part occlusion, human's orientation and pose, time range, image quality etc. Even the same person's attribute will change a lot, which make it a challenge problem.}
\label{fig:dataset_example_images}
\end{center}
\vspace{-3em}
\end{figure}

\newcommand{\tabincell}[2]{\begin{tabular}{@{}#1@{}}#2\end{tabular}}
\begin{table}[!tbp]
  \centering
  \scriptsize
  \caption{A comparison between different pedestrian attribute datasets. }
    \begin{tabular}{cccccccccc}
    \toprule
    Datasets & \#Cams & Scene & \tabincell{c}{Annotation \\ unit} & \#Samples  & Resolution & \tabincell{c}{\#Binary \\ attributes} & Viewpoint & Occlusion & \tabincell{c}{Part \\ location} \\
    \midrule
    VIPeR\cite{Gray2008reidentification} & 2 & outdoor & PID   & 1264  & $48\times128$ & 21    & yes   & no    & no \\
    PRID\cite{hirzer2011person}  & 2 & outdoor & PID   & 400  & $64\times128$ & 21    & no    & no    & no \\
    GRID\cite{liu2012person}  & 8 & outdoor & PID   & 500 & \tabincell{c}{from $29\times67$ \\ to $169\times365$}  & 21    & no    & no    & no \\
    APiS\cite{zhu2013pedestrian}  & - & outdoor & PI    & 3661  & $48\times128$ & 11    & no    & no    & no \\
    PETA\cite{deng2014pedestrian}  & - & mixture & PID   & 19,000 & \tabincell{c}{from $17\times39$ \\ to $169\times365$} &  61    & no    & no    & no \\
    \hline
    RAP   & 26 & indoor & PI    & 41,585 & \tabincell{c}{from $36\times92$ \\ to $344\times554$} & 69    & yes   & yes   & yes \\
    \bottomrule
    \end{tabular}%
  \label{tab:comparsion_with_other_datasets}%
  \vspace{-2em}
\end{table}%

Despite its importance, attribute recognition remains a challenge problem in real surveillance environments, due to the large intra-class variations in attribute categories (appearance diversity and appearance ambiguity \cite{deng2014pedestrian}). The early developed datasets, such as APiS dataset \cite{zhu2013pedestrian} and VIPeR dataset \cite{Gray2008reidentification} (annotated by Layne et al. \cite{layne2012person}), have limited training samples (several thousands of images) and sparse annotations of human attributes (dozens of binary attributes). Recently, Deng et al. \cite{deng2014pedestrian} combine 10 publicly small-scale pedestrian datasets to construct the PETA dataset, which totally consists of 19,000 images with 65 annotated attributes (61 binary attributes and 4 multiclass attriutes). Although the PETA dataset is more diverse and challenging in terms of imagery variations and complexity, most data samples in the PETA dataset are extracted from re-identification (Re-ID) datasets, which reduces the diversity of the dataset. Meanwhile, the samples are annotated based on person ID, i.e., the image samples from the same person are annotated with the same attribute set, no matter the attributes are visible or not, which is unreasonable for visual attribute detection. Furthermore, most data samples contain the whole body of pedestrian, while people are often partially visible in real surveillance scenes due to occlusion with other people or environmental objects.

Considering the above limitations, we collect a large scale Richly Annotated Pedestrian (RAP) dataset from real surveillance scenarios with 26 camera scenes and long term data collections.
Examples in the RAP dataset has shown in Figure~\ref{fig:dataset_example_images}.
The comparisons between RAP and other human attribute datasets are shown in Table~\ref{tab:comparsion_with_other_datasets}.
%Examples in RAP has been shown in Figure~\ref{fig:dataset_example_images}.
Totally, 41,585 samples are annotated with 72 fine-grained attributes (69 binary attribute and 3 multiclass attribute), which cover most attributes in description of tracking and criminal identification. It is, to the best of our knowledge, the largest human attribute dataset by far.
Moreover, to take further research on attribute recognition, three environmental and contextual factors, i.e., viewpoints, occlusion styles and body parts, are explicitly annotated.
%which provide the opportunities to quantitatively analyze their influences on attribute recognition.
On the proposed dataset, several baseline algorithms including SVM with popular hand-crafted features and convolutional neural networks (CNN) features, as well as multi-label CNN models are tested to show the challenge of attribute recognition and their performance on real world surveillance scenes.
Towards the evaluation, we also find that the commonly used attribute-independent performance metric, i.e. mean Accuracy (mA) [11], may not reflect the intrinsic dependency in multiple attributes, which is very important in multi-attribute learning.
% may not reflect the difference between variant kinds of algorithms, as the intrinsic dependency in multiple attributes are ignored by the metric.
Thus, for the first time, we propose to use the evaluation metrics in multi-label prediction \cite{zhang2014review}, i.e. accuracy, precision rate, recall rate and F1 value, as the performance measurements for attribute recognition. Extensive experimental results with variant kinds of attribute recognition algorithms validate the effectiveness of the evaluation metrics clearly.

Specially, the contributions of this paper are summarized as follows.
\begin{itemize}
\item A large-scale human attribute dataset is collected with the largest volume in data samples (41,000+) from real surveillance scenarios. More than 70 fine-grained attributes and three environmental and contextual factors (viewpoints, occlusion styles and body parts) are annotated. The dataset will facilitate the large-scale feature learning and algorithm evaluations on human attribute recognition in real scenarios.
\item A series of baseline algorithms, including SVM with the popular features and two multi-label CNN methods, are performed to show the challenges of human attribute recognition in real scenarios. Moreover, the influences of viewpoints, occlusion styles and body parts are empirically analyzed on the dataset.
\item Four evaluation metrics in multi-label learning (accuracy, precision rate, recall rate and F1 value) are firstly introduced to the benchmark evaluation. Extensive experimental results validate their effectiveness in measuring the consistency of multiple attributes predictions.
\end{itemize}

%\textcolor{red}{paper structure}
The rest of this paper is organized as follows. The next section reviews related work. Then, Section 3 describes the collected RAP dataset. Section 4 introduces the detailed evaluation methods and the baseline methods. After that, the detailed analysis about the environmental and contextual factors are discussed in Section 5 and conclusion is drawn in Section 6.

\section{Related Work}
%\textcolor{red}{fullbody based methods,lack parts}

Most of existing pedestrian attribute recognition methods are directly based on the full human body.
Layne et al. \cite{layne2012person,layne2012towards} used low-level features and Support Vector Machines (SVM) \cite{scholkopf2002learning} to detect pedestrian attributes.
They showed the great potential of attribute on person re-identification \cite{layne2012towards}.
Li et al. \cite{li2014clothing} modeled cloth attributes through the latent SVM model to assist person re-identification.
Deng et al. \cite{deng2014pedestrian} utilized the intersection kernel SVM model proposed by Maji et al. \cite{maji2008classification} to recognize attributes.
The Markov Random Field (MRF) was adopted to make a smooth on attribute prediction.
Zhu et al. introduced Gentle AdaBoost \cite{zhu2013pedestrian} to accomplish feature selection and classifier learning at the same time. Recently CNN has also been used for attribute recognition based on full body. Patrick \cite{sudowe2015person} proposed the Attribute Convolutional Net (ACN) which jointly learn different attributes through a jointly-trained holistic CNN model. The proposed model could predict multiple attributes simultaneously.
Li et al. \cite{li2015attribute} also proposed a Deep learning based Multi-attribute joint recognition model (DeepMAR) which utilized the prior knowledge in the object function for attribute recognition.

%\textcolor{red}{part and path, ignore the viewpoint and occlusion}
There is few work of attribute recognition based on part information.
Zhang et al. \cite{zhang2014panda} proposed a Pose aligned Networks for Deep Attribute modeling (PANDA).
They adopted poslets \cite{bourdev2011describing} to detect possible pedestrian parts and train an independent attribute recognition convolutional neural networks (CNN) for each part to overcome the viewpoint and occlusion problems.
Then the features from all the networks are stacked to train a linear SVM classifier for each attribute.
Differently, this paper focuses on surveillance scenarios instead of nature scene and shows quantitative results about the viewpoint, occlusion and part's influence on different attributes.
Zhu. et al. \cite{zhu2015multi} introduced the patch based Multi-label Convolutional Neural Networks (MLCNN) with predefined attribute-path connection structure to recognize attributes for person re-identification.
However the MLCNN could not cope well with the occlusion and large pose variance.

%\textcolor{red}{viewpoint and occlusion}
To our knowledge, there is no previous attempt to analyze the influence of viewpoints, occlusions and body parts on the attribute recognition in large-scale surveillance datasets. The reasons come from two aspects.
Firstly, the scale of most existing attribute datasets in surveillance scenarios is too small to analyze these factors.
Secondly, the large scale attribute datasets, such as PETA \cite{deng2014pedestrian} and APiS \cite{zhu2013pedestrian}, are heterogeneous, and lack of annotations on viewpoints, occlusions and parts. Different from existing works, this paper makes a detailed analysis about the influences of viewpoints, occlusions, body parts and interrelationships among attributes for attribute recognition on the proposed RAP dataset.
% In addition, we introduce an end-to-end CNN based multi-task learning model that considers attribute, viewpoint, occlusion and parts in a unified framework.

\renewcommand\arraystretch{1.35}
\begin{table}[!tbp]
  \centering
  \scriptsize
  \caption{Annotations in the RAP dataset.}
    \begin{tabular}{c|c|c}
    \hline
    \multicolumn{2}{c|}{Class} & Attributes \\
    \hline
    \multicolumn{2}{c|}{Spatial-Temporal} & Time, SceneID, image position,Bounding box of body/head-shoulder/upper-body/lower-body/accessories. \\
    \hline
    \multicolumn{2}{c|}{Whole} & Gender, age, body shape, role.  \\
    \hline
    \multicolumn{2}{c|}{Accessories} & Backpack, single shoulder bag, handbag, plastic bag, paper bag etc. \\
    \hline
    \multicolumn{2}{c|}{Postures,Actions} & Viewpoints, telephoning, gathering, talking, pushing, carrying etc. \\
    \hline
    \multicolumn{2}{c|}{Occlusions} & Occluded parts, occlusion types. \\
    \hline
    \multirow{3}[6]{*}{Parts} & Head & Hair style, hair color, hat, glasses. \\
          \cline{2-3}
          & Upper & Clothes style, clothes color. \\
          \cline{2-3}
          & Lower & Clothes style, clothes color, footware style, footware color. \\
          \cline{2-3}
    \hline
    \end{tabular}%
\label{tab:dataset_static}%
\vspace{-2em}
\end{table}%
\renewcommand\arraystretch{}

\section{The RAP dataset}

The RAP dataset is collected from a surveillance camera network (resolution 1,280$\times$720) at a shopping mall.
26 cameras are selected from the camera network excluding those with few pedestrians.
Considering the data richness and storage cost, totally 3 consistent months video at 15 frames per second are first collected.
After all the videos are collected, a Gaussian Mixture Model (GMM) based pedestrian detection and tracking algorithm is applied to all the videos on a ten nodes MPI cluster.
For each selected camera, one or two virtual lines are set manually indicating region-of-interest (ROI) and the tracked objects (person or accessories) accessing the virtual lines are saved automatically.
Considering the annotation cost, totally 17 hours synchronous videos are selected manually from all the collected videos for final annotation.
%Data from 17 days are selected manually due to the annotation cost.
%For each day, one consistent hour's data is sampled based on pedestrian density for final annotation.
Finally, there are 41,585 detected human images to construct the attribute dataset.
To cover most important attributes in practical surveillance system, 72 attributes are adopted in total, including some attributes that are suggested by the UK Home Office and UK police \cite{nortcliffe2011people}.
In addition, the viewpoints, occlusions, and part positions are annotated for developing better attribute recognition algorithms.

The detailed descriptions about the annotations in RAP dataset are shown in Table \ref{tab:dataset_static}.
In summary, the annotations consist six parts, including spatial-temporal information, whole body attributes, accessories, postures and actions, occlusion, and parts attributes.
To have a clear comparison with existing pedestrian attribute datasets, they are listed together in Table \ref{tab:comparsion_with_other_datasets}.
Compared to the state-of-the-art pedestrian attribute datasets, our dataset provides richer annotations and finer attributes.
In the table, PID means the annotation is based on the identification; while PI means the annotation is according to the image.
In our RAP dataset, each human image is annotated independently and the same identification person may have different attribute annotations due to the change in viewpoint, occlusion etc.
However, with PID annotations, the same attribute cross different viewpoints may perform differently, identification-based annotation may be inaccurate.
This is especially important in RAP dataset in that there are about 33 percents images are occluded.
%The PETA dataset is constructed by combining existing person re-identification datasets together.
%It has large intra-class divergence and inconsistent scene.
In addition, RAP is collected in a consistent scene and has less heterogeneity, which is desired when developing robust algorithm.
In summary, RAP contains more examples and attribute annotations than other datasets, and is therefore more suitable for studying pedestrian attributes.

Generally, the RAP dataset offers four extra four kinds of annotations compared with existing pedestrian attribute datasets, namely viewpoints, occlusions, human parts and fine attributes.
The detailed description about these annotations is as follows.
%In the following, we will give a detailed description about these four kinds of annotations.

\begin{figure}[!tbp]
\centering
\subfigure[viewpoint distribution]{
\label{subfig:viewpoint}
\begin{minipage}[t]{0.36\textwidth}
\includegraphics[width=1\textwidth]{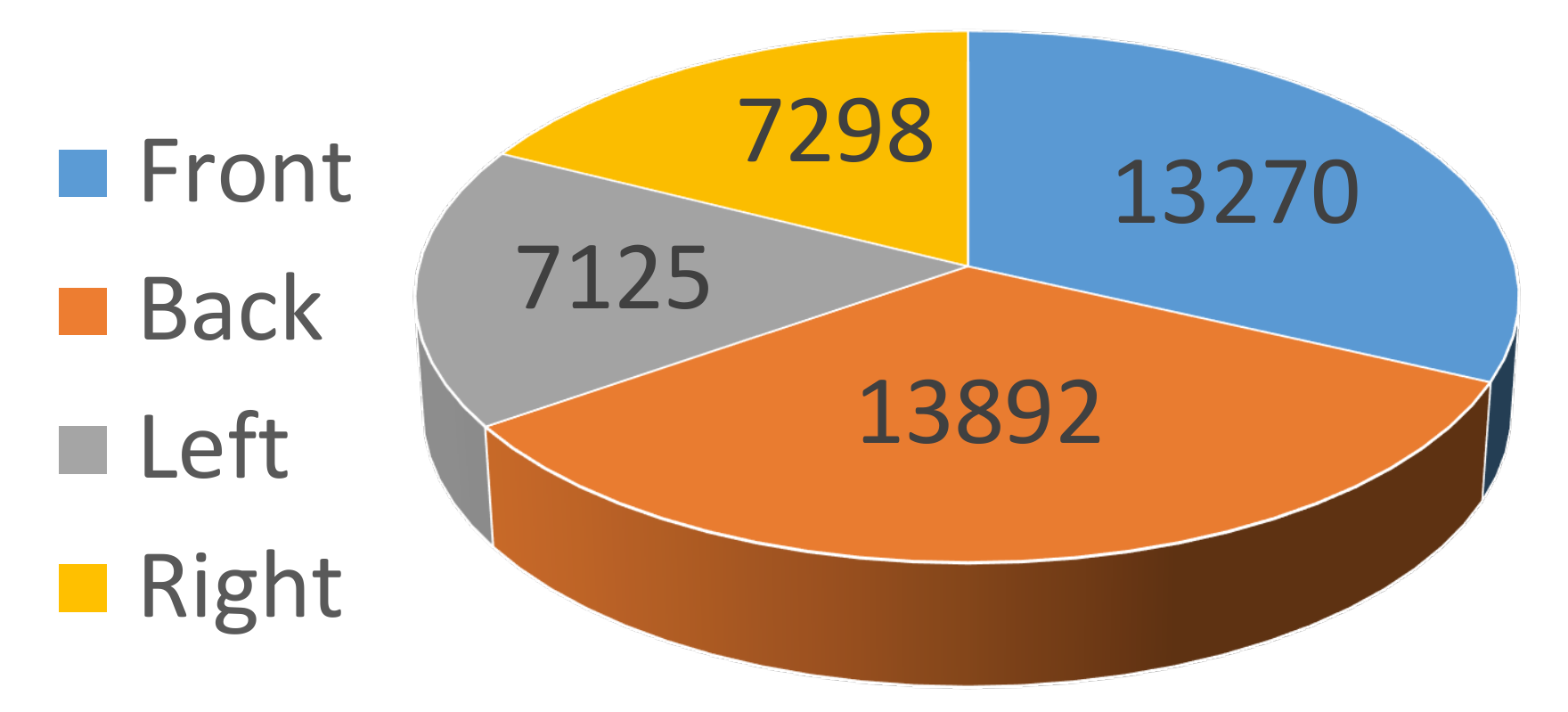}
\end{minipage}
}
\subfigure[occlusion distribution]{
\label{subfig:occlusion}
\begin{minipage}[t]{0.36\textwidth}
\includegraphics[width=1\textwidth]{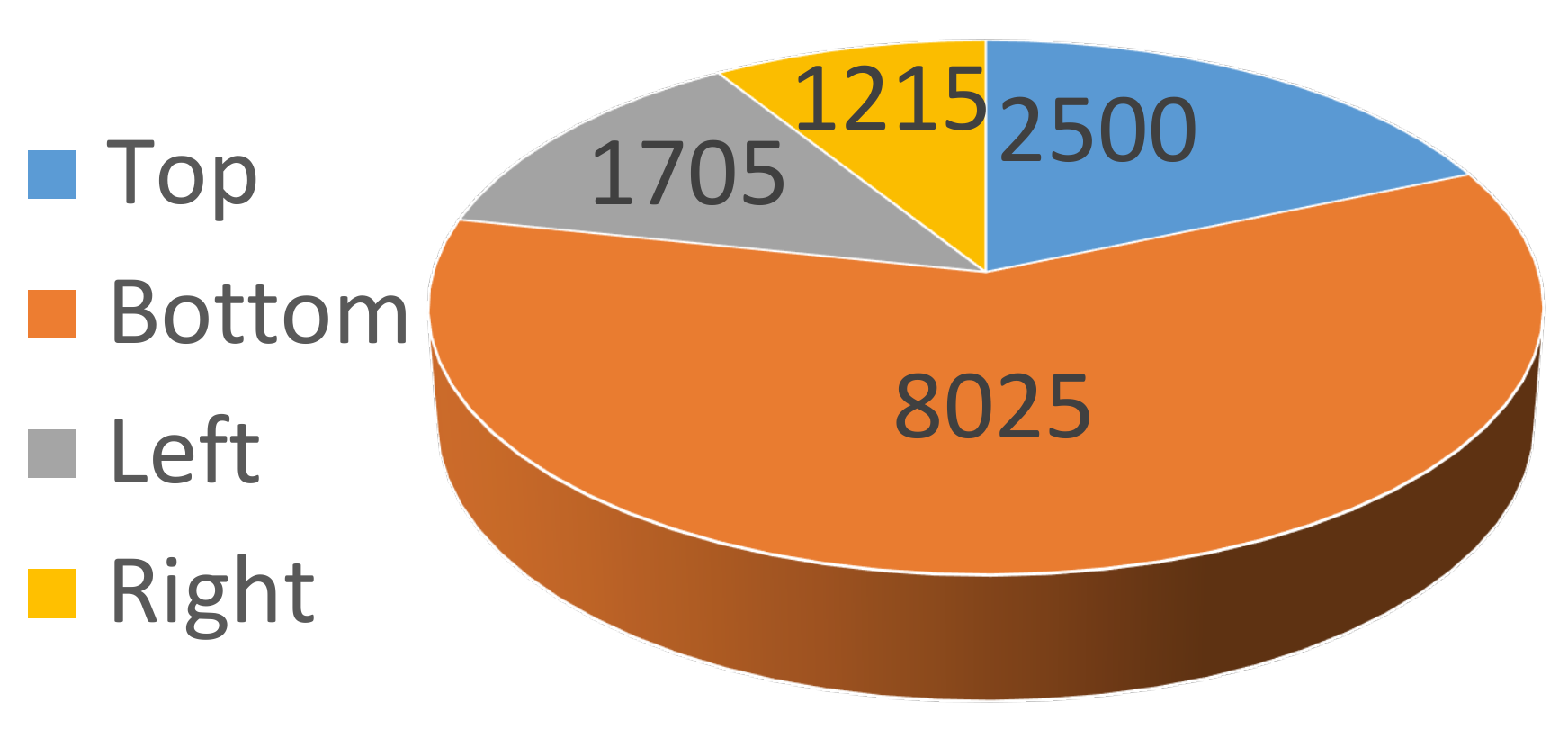}
\end{minipage}
}
\vspace{-1.8mm}\caption{(a) Distribution of different viewpoints of the annotated human images.
(b) Distribution of the annotated human images at different occlusion positions. Best viewed in color.}
\label{fig:distribution}
\vspace{-1.5em}
\end{figure}

\begin{figure}[!tbp]
\centering
\subfigure[Examples of viewpoints]{
\label{subfig:viewpoint_example_images}
\begin{minipage}[t]{0.42\textwidth}
\includegraphics[width=1\textwidth]{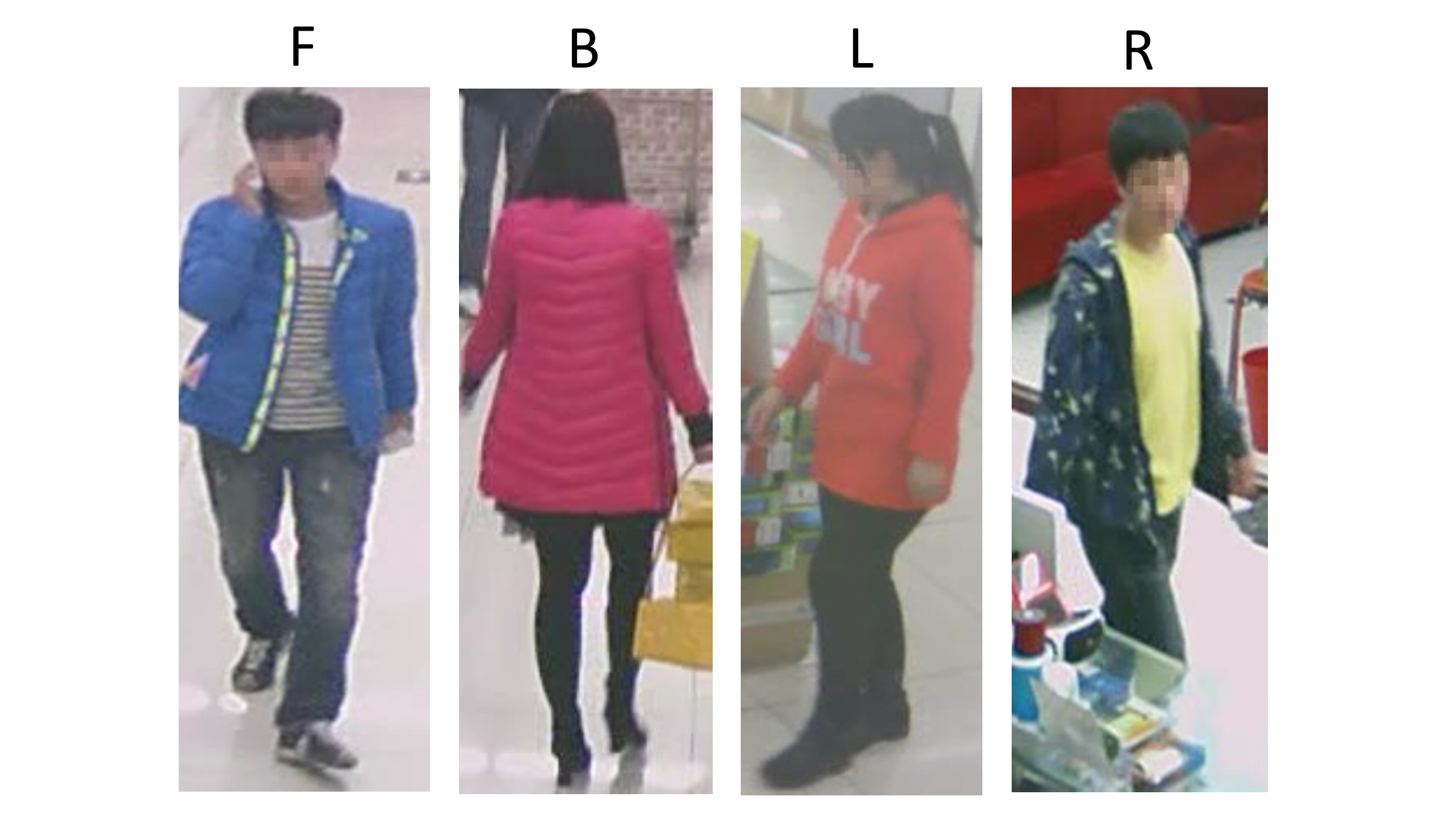}
\end{minipage}
}
\subfigure[Examples of occlusion patterns]{
\label{subfig:occlusion_example_images}
\begin{minipage}[t]{0.42\textwidth}
\includegraphics[width=1\textwidth]{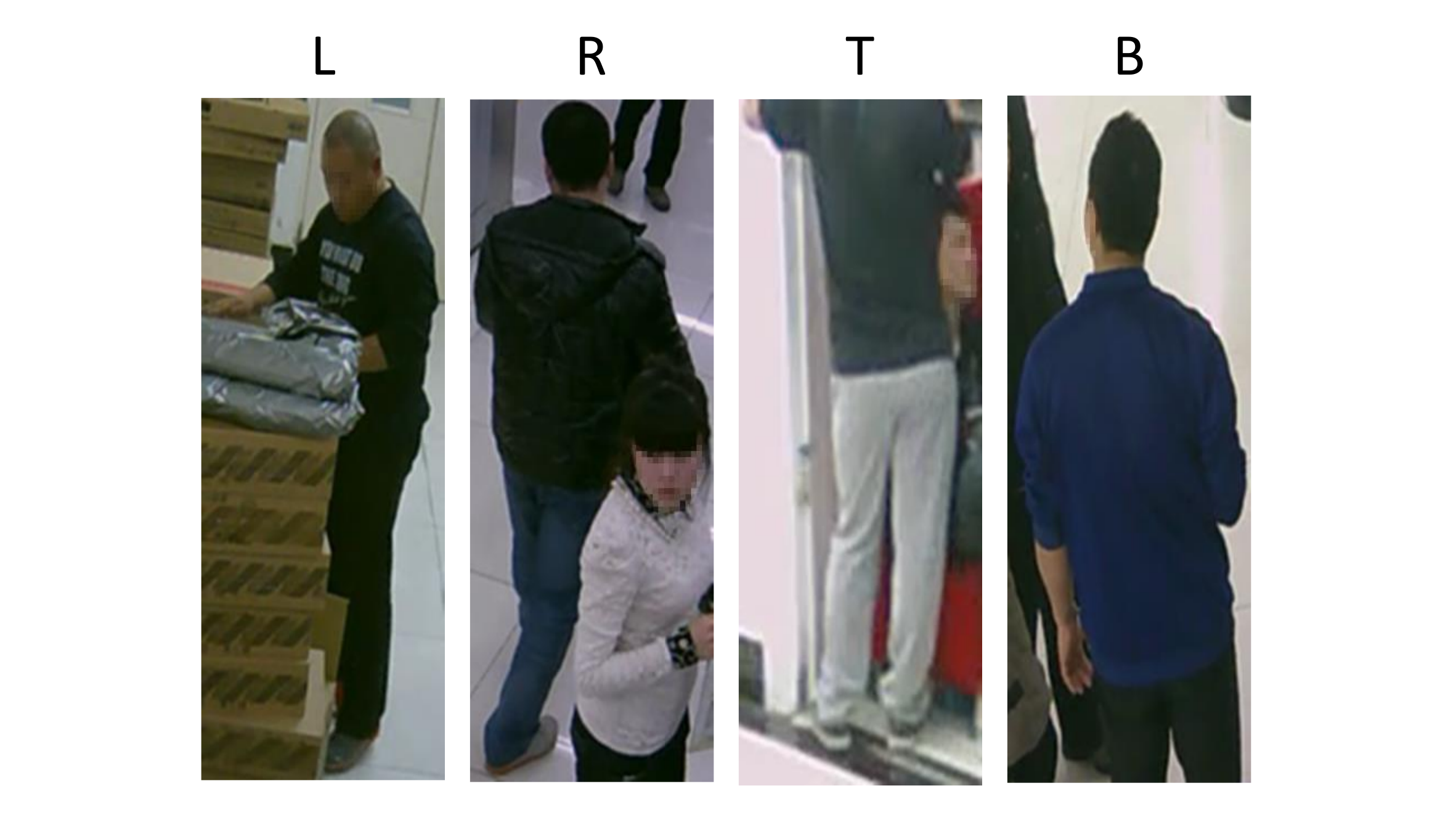}
\end{minipage}
}
\vspace{-1.5em}\caption{(a) and (b) show some examples at different viewpoints and occlusion patterns respectively. The viewpoints from left to right in (a) are front, back, left and right. The occlusion sources of first two images in (b) are attachment and person. The last two image in (b) are due to the camera border.}
\label{fig:viewpoints_occlusion_example}
\vspace{-1.5em}
\end{figure}

\textbf{Viewpoints.}
Four types of viewpoints are annotated, including facing front (F), facing back (B), facing left (L) and facing right (R).
The viewpoint is annotated based on the full body's direction relative to the annotators.
An image is not annotated if it is seriously occluded.
The distribution of different viewpoints has been shown in Figure~\ref{fig:distribution}\subref{subfig:viewpoint}.
%Viewpoints are not fully balanced as a result of the viewpoint of cameras.
Examples of different viewpoints are shown in Figure~\ref{fig:viewpoints_occlusion_example}\subref{subfig:viewpoint_example_images}.

\textbf{Occlusions.} There are four types of occlusions annotated, including top occlusion (T), bottom occlusion (B), left occlusion (L) and right occlusion (R).
An occlusion is annotated when its corresponding parts are invisible to the annotators.
The distribution of different types of occlusion is shown in Figure~\ref{fig:distribution}\subref{subfig:occlusion}.
Four sources of occlusion have also been annotated, including person, environments (such as camera border and desk), attachment (such as large package) and other.
Different from viewpoint, occlusion can be multi-value as a result of seriously occlusion by cluttered people or environment.
Example images of occlusion are shown in Figure~\ref{fig:distribution}\subref{subfig:occlusion_example_images}.

\begin{figure}[!tbp]
\centering
\subfigure[Examples of part annotation]{
\label{subfig:parts_example_images}
\begin{minipage}[t]{0.42\textwidth}
\includegraphics[width=1\textwidth]{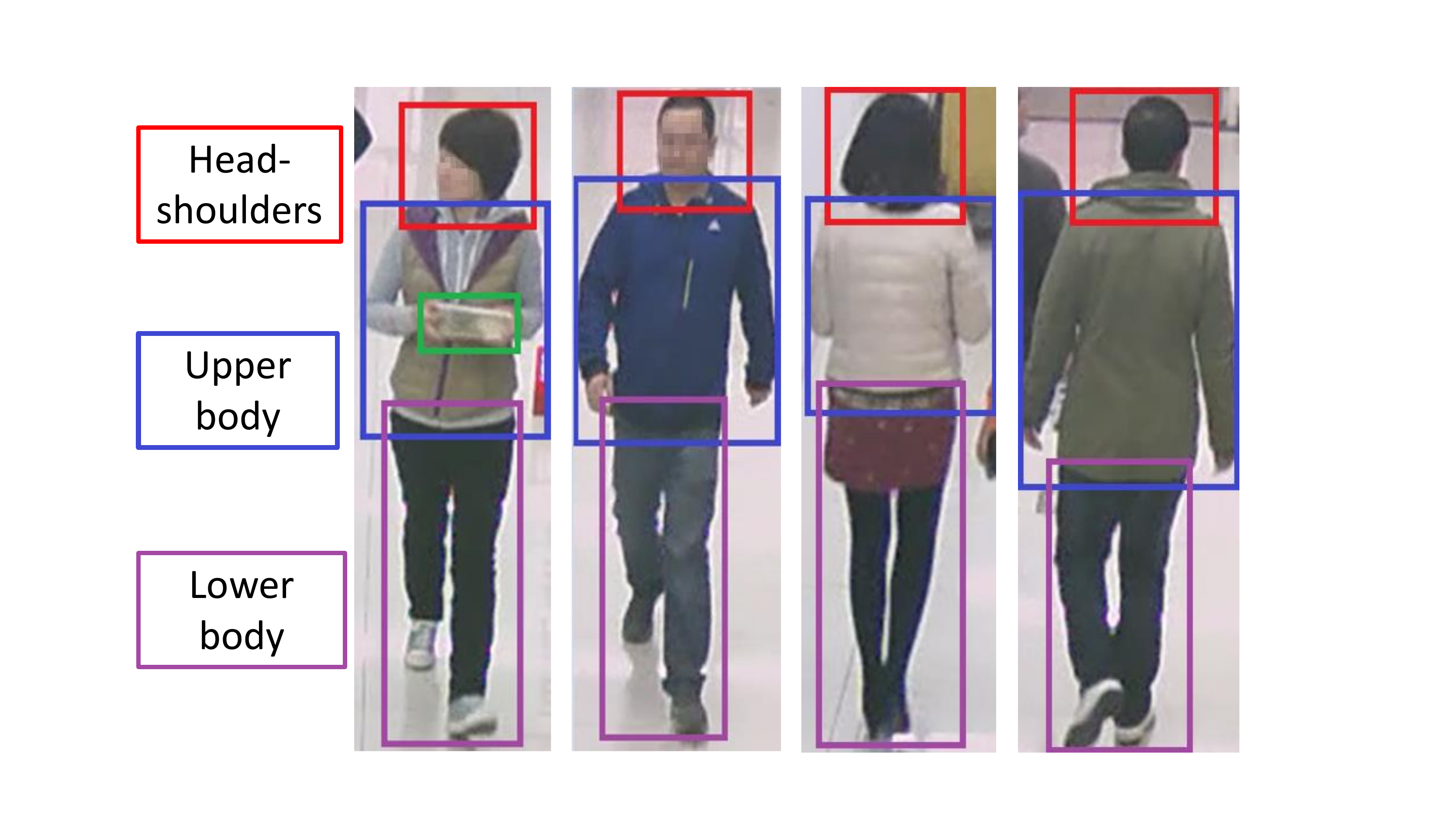}
\end{minipage}
}
\subfigure[Examples of fine attributes]{
\label{subfig:fineattribute_example_images}
\begin{minipage}[t]{0.42\textwidth}
\includegraphics[width=1\textwidth]{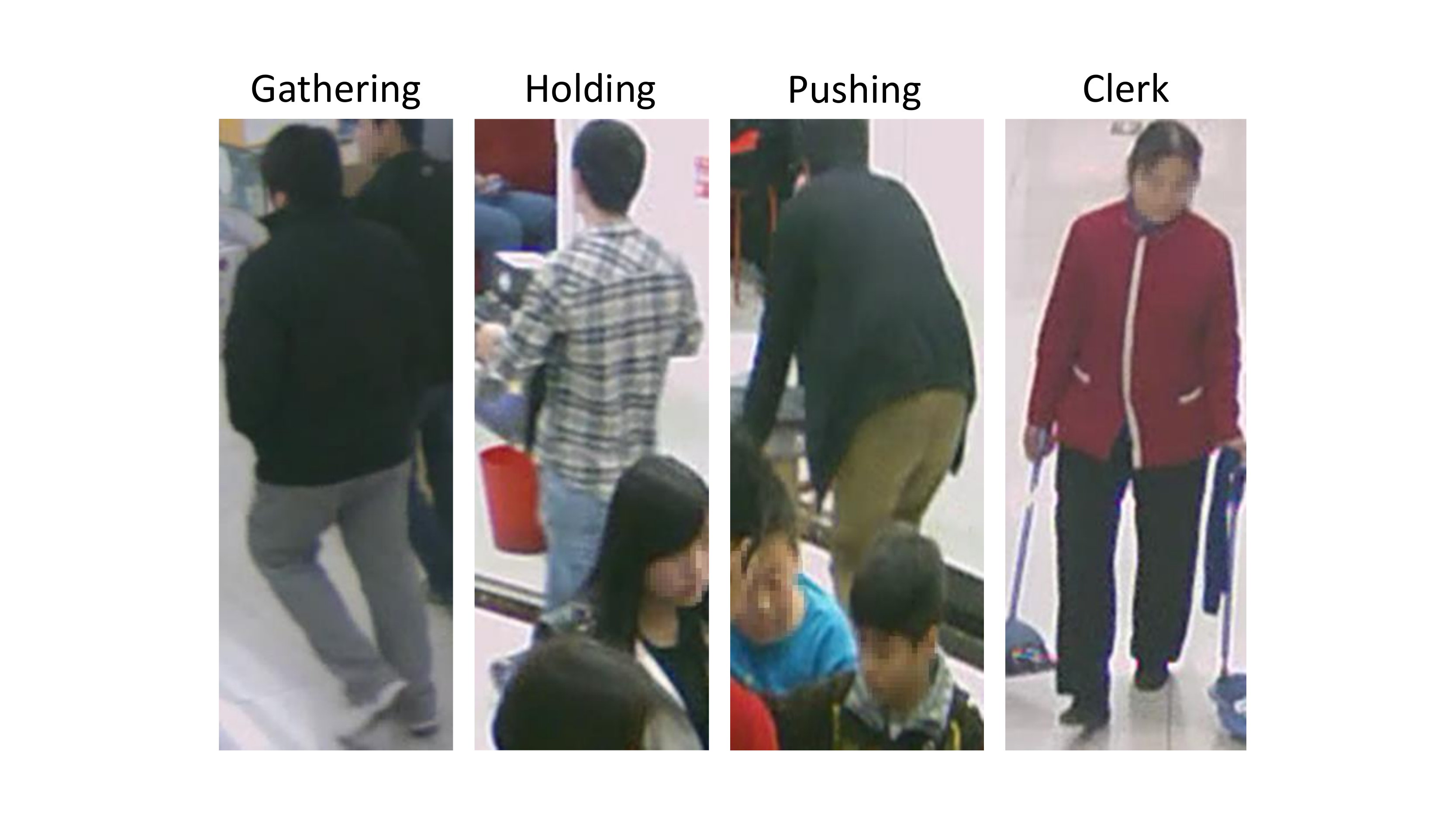}
\end{minipage}
}
\vspace{-1.5em}\caption{(a) shows examples of three parts: head-shoulder, upper body, lower body. The three parts are partly overlapped. The green bounding box in the first column is the annotation of a box. (b) shows examples of the fine-grained attributes. From left to right: gathering, carrying, pushing, major (Clerk). Best viewed in color.}
\label{fig:parts_and_fineattribute}
\vspace{-1.5em}
\end{figure}

\textbf{Body Parts.} Typically, humans in surveillance videos are relatively small and blurring.
Accurate part positions, such as for arm and face, are hard to annotate or estimate in practical systems.
Considering the above issues, instead of fine-grained part annotations, three coarse-grained parts are annotated in the RAP dataset, including head-shoulder, upper body and lower body.
These three parts link to many interesting attributes, such as hair, cloth types and shoes type.
Example of these three parts are shown in Figure~\ref{fig:parts_and_fineattribute}\subref{subfig:parts_example_images}.

\textbf{Fine Attributes.} In addition to the environmental and contextual factors above, we also labeled various interesting fine-grained attributes, including action attributes (such as talking, gathering, carrying and  pushing), role (such as customer, clerk), and body type (such as normal, fatter and thinner). Example images of these attributes are shown in Figure~\ref{fig:parts_and_fineattribute}\subref{subfig:fineattribute_example_images}.

\section{Evaluation methods and benchmark systems}
% In this section, the evaluation methods and benchmark systems will be discussed.

\subsection{Evaluation methods}
In the past, the mean accuracy (mA) is usually adopted to evaluate the attribute recognition algorithms \cite{deng2014pedestrian}.
Due to the unbalanced distribution of attributes, for each individual attribute, mA computes the classification accuracy of positive examples and negative examples separately and then take the average of them as the recognition result for the attribute.
After that, mA takes an average over all attributes as the final recognition rate.
The evaluation criterion can be formally calculated by:
\begin{equation}
\label{equ:Label_Proctol}
mA = \frac{{1}}{2N}\sum_{i = 1}^L {(TP_i/P_i + TN_i/N_i)}
\end{equation}
where $L$ is the number of attributes; $P_i$ and $TP_i$ are the numbers of positive examples and correctly predicted positive examples, respectively; $N_i$ and $TN_i$ are the numbers of negative examples and correctly predicted negative examples, respectively.

The above \emph{label-based} solution, however, treats each attribute independently and ignores the inter-attribute correlation, which exists naturally in our multi-attribute recognition problem.
For this reason, in this work we propose using the \emph{example-based} evaluation, in addition to mA.
\emph{Example-based} evaluation captures better the consistence of prediction on a given pedestrian image\cite{zhang2014review}, and it includes four metrics: accuracy, precision, recall rate and F1 value, as defined below:
% However, the pedestrian attribute recognition problem is a multi-label learning problem. The label-based evaluation can not reflect the relationships among attributes for the same human.
%Compared with the label-based evaluation (eq. Formula~\ref{equ:Label_Proctol}), the example-based evaluation may better capture the consistence of prediction on a given pedestrian image\cite{zhang2014review}. In this paper, despite the label based evaluation, the example-based evaluation method is also introduced to evaluate the attribute classification results, including accuracy, precision, recall rate and F1 value.

\begin{equation}
\label{equ:Instance_Proctol}
\renewcommand\arraystretch{2.0}
\begin{array}{r}
Acc_{exam} = \frac{1}{N}\displaystyle\sum_{i=1}^N \frac{|Y_i\bigcap f(x_i)|}{ |Y_i \bigcup f(x_i)|},\quad Prec_{exam} = \frac{1}{N}\displaystyle\sum_{i=1}^N \frac{|Y_i\bigcap f(x_i)|}{ |f(x_i)| } \\
Rec_{exam} = \frac{1}{N}\displaystyle\sum_{i=1}^N \frac{|Y_i\bigcap f(x_i)|}{ |Y_i| }, \quad F1 = \frac{2*Prec_{exam}*Rec_{exam}}{Prec_{exam} + Rec_{exam}}
\end{array}
\end{equation}
where $N$ is the number of examples, $Y_i$ is the ground truth positive labels of the $i'th$ example, $f(x)$ returns the predicted positive labels for $i'th$ example.\; and $|\cdot|$ means the set cardinality.

\subsection{Benchmark systems}
\textbf{Baseline methods.} %As a popular approach in machine learning, the SVM approach has obtained great successes in many vision related topics, such as object classification and object detection.
Previous work \cite{layne2012person,layne2012towards} has shown that SVM can be utilized to recognize pedestrian attributes.
In this paper, we adopt \textbf{SVM} algorithm as one of the baseline attribute recognition algorithms.
For implementation, the liblinear package\footnote{ \url{http://www.csie.ntu.edu.tw/~cjlin/liblinear/} } is applied for the task.
Different from previous work \cite{layne2012person,layne2012towards,deng2014pedestrian}, the linear kernel is used instead of intersection kernel in this paper due to similar result in performance.
The parameter C is selected through cross validation in \{0.01, 0.1, 1, 10, 100\}.
In our experiments, we find that the model trained on the ELF feature achieves best results when C is 1, while the model trained on the CNN features performs the best when C is 0.1. Generally, attribute usually has unbalanced distribution.
In the training stage, not all the training examples are selected for training the linear SVM model.
Similar as Layne et al. \cite{layne2012person}, we randomly down sample the negative examples to the size of positive samples if positive samples is less than negative samples. The same rules apply to the positive samples.
The sampled subset is used for training the linear SVM model.
All the test data are used for test.

For the multi-attribute joint learning, we implement two CNN models based on Caffe \cite{jia2014caffe} framework, including the \textbf{ACN} \cite{sudowe2015person} and \textbf{DeepMAR} \cite{li2015attribute}. The ACN has achieved well performance in Parse-27k \cite{sudowe2015person} and Berkeley Attributes of People \cite{bourdev2011describing}.
The DeepMAR has achieved current state-of-the-art performance in the PETA dataset.
Both of ACN and DeepMAR adopted CaffeNet\cite{jia2014caffe}, which is based on the Alexnet \cite{krizhevsky2012imagenet}, and add extra fully connected layers for each attribute or use different loss functions.
The network structure of ACN and DeepMAR in this paper is the same to the initial papers.
For the DeepMAR model, the initial learning rate is 0.001 and the weight decay is 0.005.
For the ACN model, the initial learning rate is 0.001 and the weight decay is 0.05.
The learning rate will divide 10 after 2000 iterations with batchsize 256.
Both of these two models adopt SDG solver to train.
For data augmentation, the mirror and random crop from 256*256 to 227*227 is adopted.
The model at 15,000 iterations is adopted for the final evaluation for ACN and DeepMAR.

\textbf{Features.}
We adopt two types of features as for the SVM model, including Ensemble of Localized Features (\textbf{ELF}) feature and CNN feature.
The ELF feature, which is proposed by Gray et al. \cite{Gray2008reidentification} and later modified by Prosser et al. \cite{prosser2010person}, has been successfully used in human attribute recognition \cite{layne2012person,deng2014pedestrian,layne2012towards}.
It consists of 8 colour channels (RGB, HS, and YCbCr) and 21 texture filters (Gabor \cite{fogel1989gabor} and Schmid \cite{schmid2001constructing}) derived from the luminance channel.
We adopt ELF as the baseline feature, using the same parameter setting for $\gamma, \lambda, \theta, \sigma$ as in~\cite{prosser2010person} for Gabor filter extraction, and same $\tau$ and $\sigma$ for the feature in~\cite{schmid2001constructing}.
Lastly, a 16 bin feature is extracted for each channel.
As the same with existing work\cite{Gray2008reidentification,prosser2010person}, we divided the human image into six equal strips.
For each strip, a 474 dimensional ELF feature is extracted.
In the last, totally 2784 dimensional low-level feature vector is created to represent the image.
The open source code implement in github\footnote {\url{https://github.com/rlayne/ELF-v2.0-Descriptor}} is introduced to extract feature.

Recently, CNN has been successfully used in vision related areas, such as object recognition \cite{krizhevsky2012imagenet}, object detection \cite{girshick2014rich} and face recognition \cite{taigman2014deepface}.
The features generated by a CNN model which is trained for large-scale object recognition tasks have also show a good generalization ability among general object recognition tasks\cite{donahue2013decaf,razavian2014cnn}, such as object image classification, fine-grained recognition, attribute detection.
Besides the ELF feature, the CNN feature is also adopted as baseline features in this paper.
We adopt BVLC Reference CaffeNet\footnote {\url{http://caffe.berkeleyvision.org/model\_zoo.html}}, which is trained on the 1.2M images from the
ImageNet ILSVRC-2012 1000-way classification task, to generate the CNN features in our following experiments.
The \textbf{FC6} and \textbf{FC7} layers are used to extract features, both of which have 4096 dimensions and are commonly used in recognition tasks.
Notably, both of these extracted features have been L2 normalized for the SVM model.

\section{Experiments}
In this section, the experiment setup will be introduced first.
Then, the overall experiment results will be described.
Finally, the detailed analysis about the four environmental and contextual factors' (viewpoints, occlusions, body parts, interrelationships among attributes) influence on attribute recognition tasks will be discussed.

\textbf{Setup.}
The experiments are conducted with 5 random splits.
For each split, totally 33,268 images are used for training and the rest 8,317 images are used for test.
Due to the highly unbalanced distribution of attributes, totally 51 binary attributes are selected in our experiments.
The attribute is selected when the positive examples' ratio in the dataset is higher than 0.01.
In addition, the baldhead is also selected due to its salience in real scenario.
To have a fair analysis of each factors, the human images are divided into two classes: occlusion images and clean images.
The detailed information about how to use these two class data is in the following subsections.

\begin{figure}[!tbp]
\begin{center}
\includegraphics[width=1.0\linewidth]{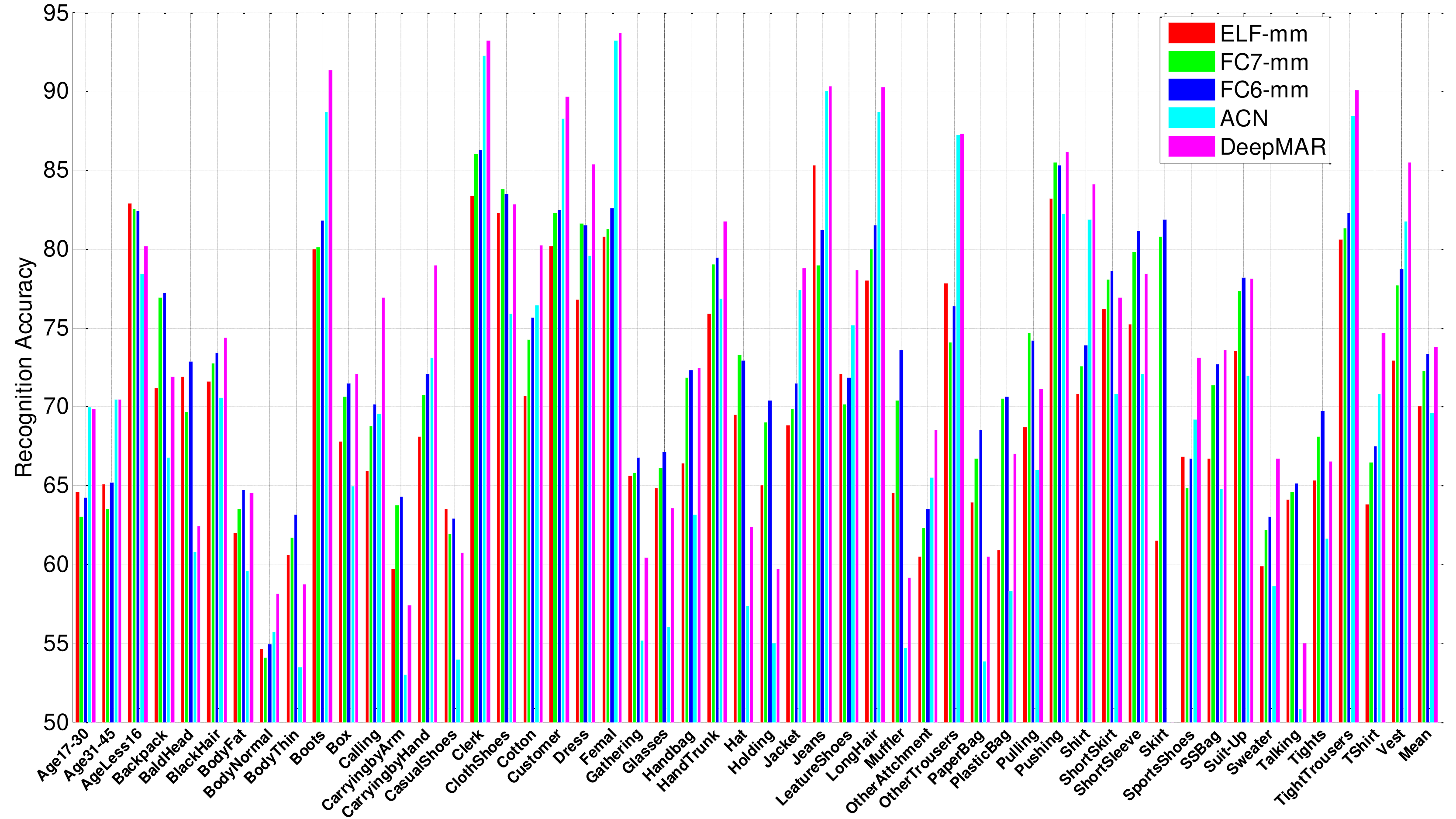}
\vspace{-3em}
\end{center}
\caption{Label based attribute recognition results on RAP(\%). Best viewed in color.}
\label{fig:baselineResult}
\vspace{-2.0em}
\end{figure}

\subsection{Overall evaluation}
\label{subsection:baseline}

This paragraph gives a general recognition result without distinguishing the clean and occlusion images in training and test stages.
Considering each attribute independently, a linear SVM model is trained for each attribute through features ELF, FC6 and FC7.
The results have been shown in Figure~\ref{fig:baselineResult}.
The first and second character ``m" in ELF-mm represent the data composition in the training and test stage separately.
The characters ``o",``c",``m" represent occlusion data only, clean data only and the mixture of occlusion and clean data respectively.

Obviously, the attributes for gender, backpack, role, trousers and action of pushing have better recognition accuracy.
This is consistent with Deng et al. \cite{deng2014pedestrian}.
Compared with the ELF feature, the CNN feature trained at the large-scale object recognition tasks has shown better performance in human attribute recognition task than ELF feature.
Compared with the FC7 layer feature, the FC6 layer feature has better generalization, which is consistent with Donahue et al. \cite{donahue2013decaf}.
So in the following subsections, the FC6 layer feature is the default feature if there is no special notices.
For the example based evaluation, we will show the results on Table \ref{tab:accuracy_variant_evaluation} for further comparison.

\subsection{Factors analysis}
In this subsection, the detailed analysis about four environmental and contextual factors: viewpoints, occlusions, parts, and the relationship among attributes will be discussed.
%At the end, the proposed CNN based multi-attribute jointly recognition baseline is explained.

\begin{figure}[!tbp]
\centering
\subfigure[Recognition results at different viewpoints]{
\label{subfig:result_viewpoint}
\begin{minipage}[t]{0.48\textwidth}
\includegraphics[width=1\textwidth]{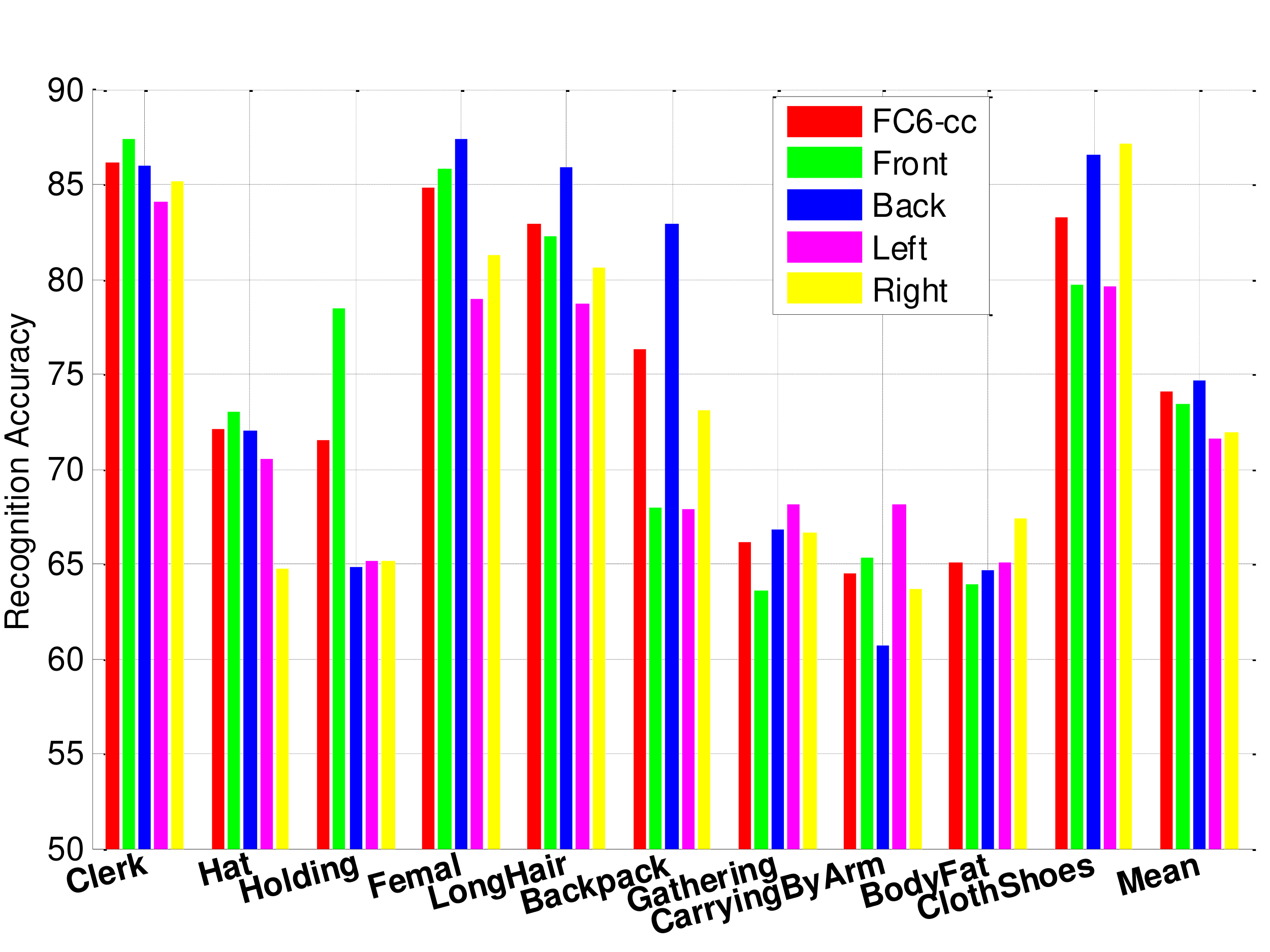}
\end{minipage}
}
\subfigure[Examples of recognition results.]{
\label{subfig:viewpoint_recognition_results}
\begin{minipage}[t]{0.48\textwidth}
\includegraphics[width=1\textwidth]{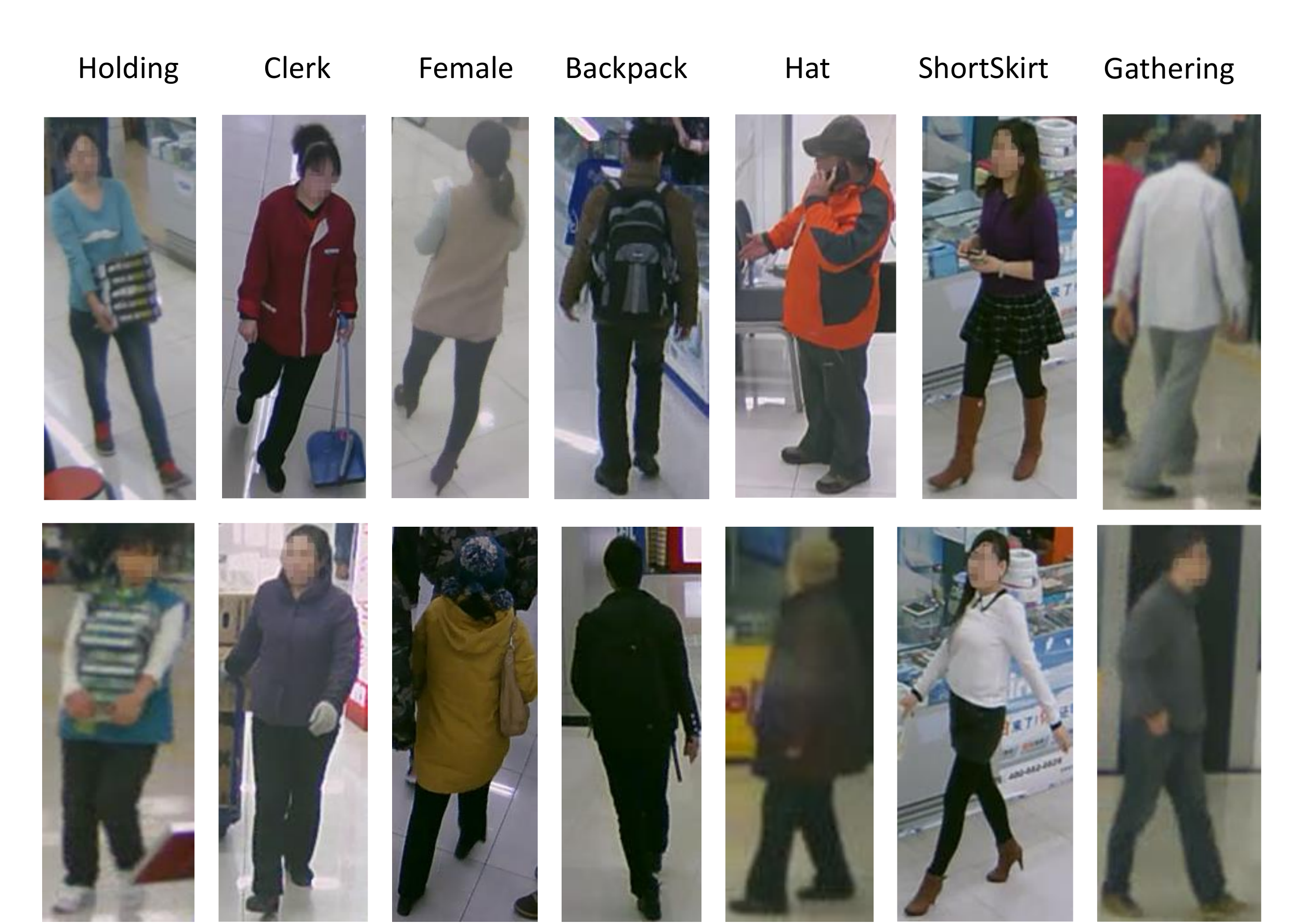}
\end{minipage}
}
\vspace{-1.8mm}\caption{Figure (a) shows some attributes' recognition accuracy at different viewpoints. Figure(b) gives some corresponding recognition results. The first row is true positives and the second row is false positives. Best viewed in color.}
\label{fig:viewpoint_recognition}
\vspace{-2em}
\end{figure}

\textbf{Viewpoint} As a common mentioned factor in vision related topics, viewpoint has great influence in many tasks.
A qualitative analysis about the influence of viewpoint on attribute recognition will be discussed.

To better analyze the influence of viewpoint, the occlusion human images are eliminated and only the clean data are selected in both training and test stage.
So the influence of occlusion factor can be removed.
In the training stage, all the clean data with different viewpoints are feeded into the linear SVM model.
In the test stage, all the test examples are divided into four parts based on the four groundtruth viewpoints, including front, back, left and right.
Each attribute will get a mean recognition accuracy at each viewpoint.
For clear comparison, the average recognition accuracy of all the clean test examples is also presented, which is named as FC6-cc that are trained and tested using clean data based on FC6 layer feature.
The recognition results of partial selected attributes and the average recognition of all the attributes in different viewpoints have shown in Figure~\ref{fig:viewpoint_recognition}\subref{subfig:result_viewpoint}.

Generally, some attributes are easily recognized in certain viewpoints.
This is obvious as shown in Figure~\ref{fig:viewpoint_recognition}\subref{subfig:result_viewpoint}.
Attributes, such as Cleark, Holding and Hat, are very easy to be recognized at front viewpoint.
This is easy to understand because these attributes are better viewed in front viewpoint.
Recognizing clerk in front viewpoint is easy in that the uniform is special in front viewpoint.
LongHair is also easy to be recognized because many women could like to put their hair in the back.
Besides, people also would like put their backpack on their back.
When people carries something in the upperarm, it is easy to recognize from the left and right viewpoint.
From the last row in Figure~\ref{fig:viewpoint_recognition}\subref{subfig:result_viewpoint}, it could be concluded that attribute classification in front and back viewpoints is relatively easier than in left and right viewpoints.
To better understand the results, some attributes recognition results at different viewpoints have been shown in Figure~\ref{fig:viewpoint_recognition}\subref{subfig:viewpoint_recognition_results}.
The top row shows the attributes that has been classified correctly and the bottom row shows the corresponding wrongly classified attributes.
Each column represents an attribute.

As we can see from Figure~\ref{fig:viewpoint_recognition}\subref{subfig:result_viewpoint}, viewpoint plays an important role in attribute recognition.
With viewpoint prior, many vision tasks could be better solved such as object detection \cite{felzenszwalb2010object}.
So classifying the viewpoint is also an important task.
Treating each viewpoint as an independent attribute, like other attributes, viewpoint could be classified using above attribute recognition methods.
The classification results have been shown in Table \ref{tab:viewpoint_classification}.
% The character ``m" and ``c" are the same as in \ref{subsection:baseline}.
As shown in Table \ref{tab:viewpoint_classification}, the FC6 layer feature obtains the best results in mixture data in all the four types of viewpoints.
After removing the occlusion data in training and test stage, the viewpoint classification accuracy is improved by 1.2\%, which shows promising potential in handing viewpoint problem.

% Table generated by Excel2LaTeX from sheet 'Sheet2'
\begin{table}[!tbp]
\vspace{-2em}
\centering
  %\scriptsize
  \caption{Viewpoint classification results. The bold number is the max value each row.}
    \begin{tabular}{ccccc}
    \toprule
    Viewpoint & \quad FC6-cc & \quad ELF-mm & \quad FC7-mm & \quad FC6-mm \\
    \midrule
    FacingFront & \textbf{89.42}  & 84.47  & 86.51  & 87.35  \\
    FacingBack  & \textbf{90.09}  & 84.67  & 87.41  & 88.14  \\
    FacingLeft & \textbf{84.92}  & 75.03  & 80.98  & 83.94  \\
    FacingRight & \textbf{85.74}  & 74.81  & 80.34  & 83.96  \\
    \hline
    Mean  & \textbf{87.54} & 79.74 & 83.81 & 85.85 \\
    \bottomrule
    \end{tabular}%
\label{tab:viewpoint_classification}%
\vspace{-2em}
\end{table}%

\textbf{Occlusion} Occlusion is an important variable in many visual recognition tasks.
To handle the occlusion problem, few existing work focuses on detecting effective parts to assist object recognition, object detection.
However, much existing work ignores this factor due to some reasons.
One of the reasons may be that it is hard to model occlusion explicitly in object functions.
To our knowledge, no existing work gives a quantitative experiment analysis about the influence of occlusion.
In this part, we will give a detailed analysis about this factor on attribute recognition tasks.

\begin{figure}[!tbp]
\begin{center}
\includegraphics[width=0.7\linewidth]{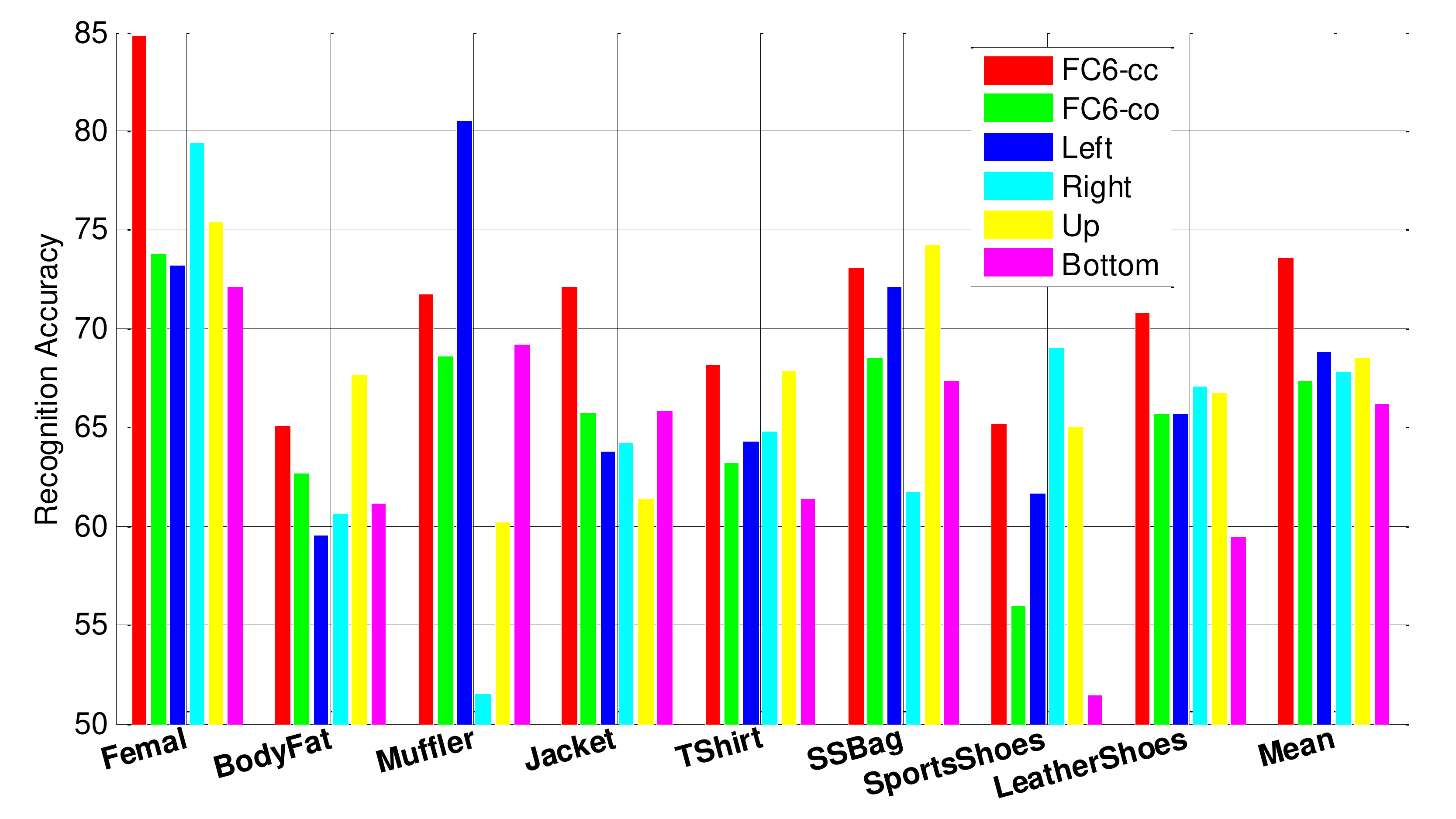}
\vspace{-1.8mm}
\caption{Partial attribute recognition results of occlusion images.}
\label{tab:result_occlusion}
\end{center}
\vspace{-3em}
\end{figure}

In order to analyze the influence of the occlusion factor, the clean data in training set are used in the training stage in this part.
Only the occlusion data in the test set are used for test (FC6-co in Figure~\ref{tab:result_occlusion}).
For comparison, the results using clean test data are also shown.
%Because the occlusion distribution is little seriously unbalanced, some attributes, such as BaldHead, Hat, and Muffler, have no examples in the test stage at partial occlusion positions.
%To have a fair comparison, these attributes who have no test examples in any occlusion positions are excluded when computing the mean recognition accuracy.
This rule is applied to both the FC6-cc and FC6-co.
The partial attributes' recognition results are shown in Figure ~\ref{tab:result_occlusion}.
%The selected attributes are divided into four groups from top to bottom, including full body, head-shoulder, upper body and lower body related attributes.
As shown in the Figure ~\ref{tab:result_occlusion}, compared with the results of FC6-cc (clean data are used in training and test stage), near all the attributes' recognition accuracies drop a lot due to occlusion, such as gender, SSBag (Single Shoulder Bag) which is located at upper body, and LeatherShoes and SportsShoes which is located at lower body.
The average accuracy of all the attributes drops 5.6\%.
However, results of attributes like BodyFat which corresponds with full body only change a little.
This shows that the attributes which are corresponded with parts are more easily affected by occlusion while attributes are corresponded with full body are more robust.

%% Table generated by Excel2LaTeX from sheet 'Sheet11'
%\begin{table}[!tbp]
%  \centering
%  %\scriptsize
%  \caption{Partial attribute recognition results of occlusion images. The bold number is the max value in each row of the right six
%columns.}
%    \begin{tabular}{ccccccc}
%    \toprule
%    Attribute & FC6-cc & FC6-co & Left     & Right     & Up     & Bottom \\
%    \midrule
%    Femal & \textbf{84.83 } & 73.79  & 73.17  & 79.38  & 75.42  & 72.12  \\
%    BodyFat & 65.12  & 62.66  & 59.51  & 60.63  & \textbf{67.65 } & 61.16  \\
%    Muffler & \textbf{71.76 } & 68.56  & 80.51  & 51.49  & 60.22  & 69.16  \\
%    Jacket & \textbf{72.13 } & 65.71  & 63.79  & 64.24  & 61.39  & 65.83  \\
%    T-Shirt & \textbf{68.18 } & 63.16  & 64.29  & 64.80  & 67.84  & 61.34  \\
%    SSBag & 73.08  & 68.50  & 72.09  & 61.71  & \textbf{74.19 } & 67.32  \\
%    SportsShoes & 65.13  & 55.92  & 61.66  & \textbf{69.02 } & 65.03  & 51.45  \\
%    LeatherShoes & \textbf{70.78 } & 65.68  & 65.70  & 67.03  & 66.79  & 59.46  \\
%    \hline
%    Mean  & \textbf{73.55 } & 67.33  & 68.79  & 67.79  & 68.55  & 66.21  \\
%    \bottomrule
%    \end{tabular}%
%\label{tab:result_occlusion}%
%\vspace{-2em}
%\end{table}%

% to show the example results on different parts.
\textbf{Part position} Part plays an important role in many vision topics, such as object detection and  fine-grained object recognition.
Some existing work \cite{zhang2014panda,zhu2015multi} also utilized part or patch to assist human attribute recognition.
They mainly focused on the ensemble of different parts' information or pre-defined structure between attribute and parts based on the human's concept to assist recognition of attributes.
In this discussion, human attributes recognition results only based on parts information will be displayed.

\begin{figure}[!tbp]
\centering
\subfigure[Recognition results at different parts]{
\label{subfig:result_parts}
\begin{minipage}[t]{0.48\textwidth}
\includegraphics[width=1\textwidth]{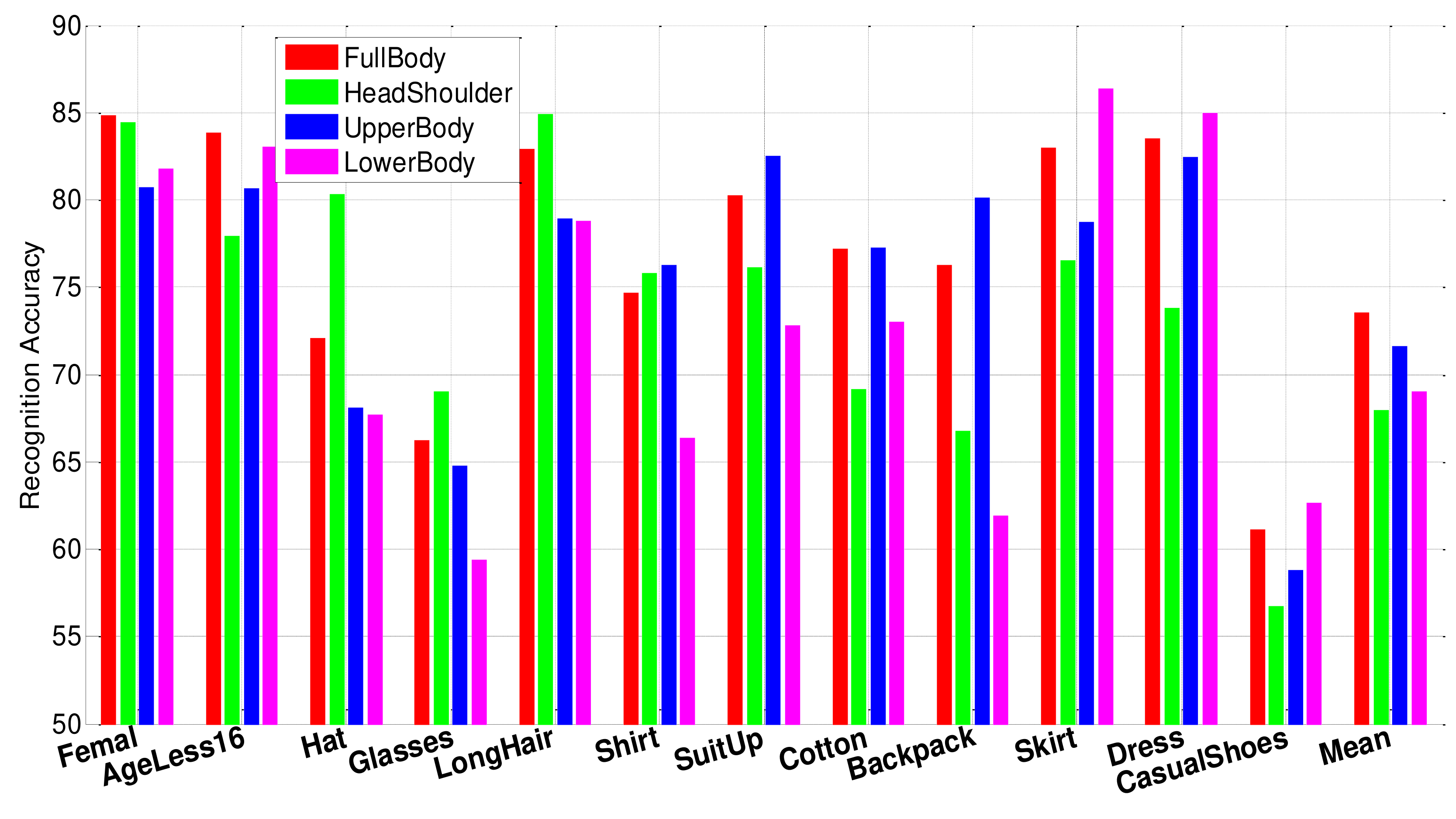}
\end{minipage}
}
\subfigure[Examples of recognition results.]{
\label{subfig:upbody_recognition_result}
\begin{minipage}[t]{0.48\textwidth}
\includegraphics[width=1\textwidth]{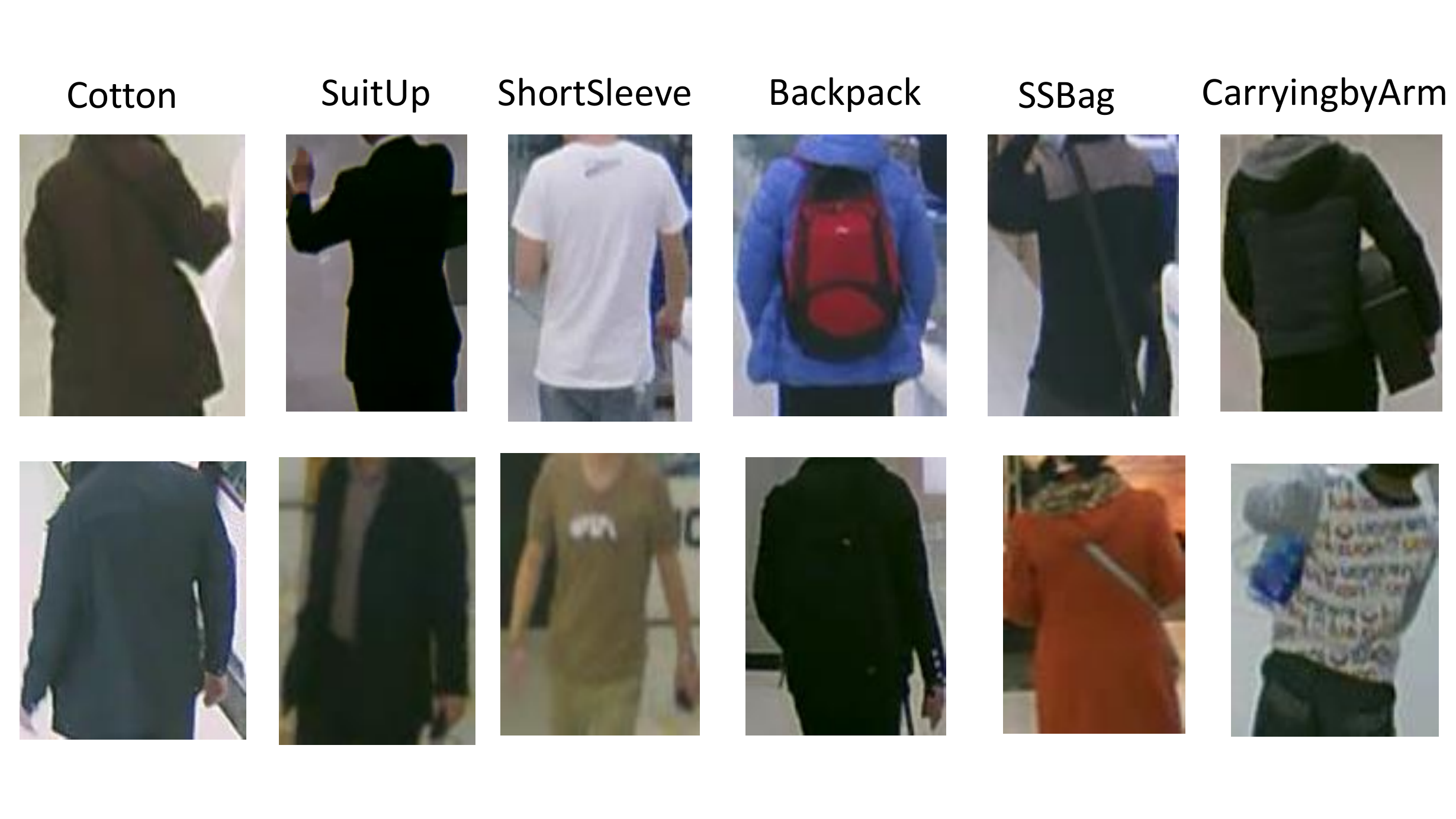}
\end{minipage}
}
\vspace{-1.8mm}\caption{Figure (a) shows some attributes' recognition accuracy using part image. Figure(b) gives some corresponding recognition results at using UpperBody. The first row is true positives and the second row is false positives. Best viewed in color.}
\label{fig:parts_recognition}
\vspace{-2.5em}
\end{figure}

To analyze the part's influence on human attribute recognition task, only the clean human images without occlusions are selected in training and test stage.
For each selected human image, the predefined three parts including head-shoulder, upper-body, lower-body are extracted based on the ground truth position information.
For each part, a linear SVM model is trained for all selected attributes.
In the test stage, the ground truth part images are used to recognize attributes.
This could reduce the influence caused by the error of part detection.
The overall recognition results of selected attributes are shown in Figure~\ref{fig:parts_recognition}\subref{subfig:result_parts}.
The value of FullBody represents the mean recognition accuracy of FullBody without using part information.

From the perspective of human, some attributes are strongly relevant with fixed parts.
For example, longhair and glasses depend only on the head-shoulder, while short skirt exist only on the lowerbody part.
This is also consistent with our experiments results in Figure~\ref{fig:parts_recognition}\subref{subfig:result_parts}.
To have a better view about the recognition results, the selected attributes are divided into groups based on the highest recognition results according parts.
The groups includes four types: fullbody related attributes, header-shoulder related attributes, upperbody related attributes and lowerbody related attributes.
For example, the BlackHair and Glasses obtain highest recognition accuracy based on the HeadShoulder part, so they are assigned to the headshoulder related attribute group.
Finally, the attribute are sorted by groups with the order fullbody, head-shoulder, upperbody and lowerbody related attributes.

As shown in Figure~\ref{fig:parts_recognition}\subref{subfig:result_parts}, the highest accuracy of Female is obtained using Fullbody features.
It means that the gender is a global attribute which corresponds to more than one part.
% For the attribute LongHair, the high accuracy is achieved only using HeadShoulder part features.
The LongHair attribute achieves high accuracy using HeadShoulder part features, which means it is highly related with HeadShoulder part.
% It shows this attributes is highly related with HeaderShoulder part.
This is consistent with human's concept that hair locates at HeaderShoulder.
The similar experimental results can be seen on the other two groups.
For example, Suit-up and Backpack are easily recognized using the UpperBody features,
thus they own high correlations with UpperBody part.

In summary, recognizing attributes in upperbody is easier than two parts.
This is relatively clear in that rich information exists in UpperBody part.
Some attribute results based on UpperBody have been shown in Figure~\ref{fig:parts_recognition}\subref{subfig:upbody_recognition_result}.
Attribute recognition according to parts is relative easier than just using full body information.
Using information, such as parts, to guide pedestrian attribute recognition is an promising direction.

\textbf{Relationship among attributes}
\label{relationship}
Some attribute are corresponded with each other. Learning the attributes classifier simultaneously could be better to capture the relationship among attributes than learn each attribute alone.
In this part, we will give the results using ACN model and DeepMAR model, and both of them could learn all the attribute simultaneously.

The overall multi-attribute joint learning results has been shown on Table \ref{tab:accuracy_variant_evaluation}.
For clear comparison, the results of SVM using three types of features are shown using \emph{example-based} evaluation.
In the table, through joint learning all the attributes simultaneously, the performance using \emph{label-based} evaluation could improve little or even drops on RAP dataset using ACN and DeepMAR. The reasons may come from the extremely unbalance distribution of single attribute. We make a static about the results' distribution on RAP dataset and find that for the attributes that the DeepMAR model has lower performance than other methods(a total of 22), the positive examples ratio that are less than 0.03 could reach to 70 percents(a total of 16). It is necessary to train a deep model with enough examples. For the SVM, the positive examples and negative examples are sampled to be equal.
This training strategy helps the SVM to have better performance than DeepMAR on some extremely unbalance distribution attributes.

Different with the \emph{label-based} evaluation, both the ACN and DeepMAR can obtain great improvements using \emph{example-based} evaluation. Especially, the multi-attribute joint learning model could improve more than 25 percents than the SVM model using \emph{example-based} evaluation, including Accuracy and Precision.
It shows better ability in the predictions' consistence on a pedestrian image.
This is very useful for real application.
In real application, people will search a person using a series of attributes.
The prediction's consistence is very important for attribute based searching.

To verify the baseline methods' efficiency, we take further experiments on PETA dataset, which has been shown on Table \ref{tab:accuracy_variant_evaluation}. The features used in PETA for SVM model are the same to RAP. The experiment setting is the same as the paper \cite{deng2014pedestrian}. Different with RAP dataset, the \emph{label-based} evaluation mA using ACN and DeepMAR could be higher than SVM in PETA dataset. This owns to the balance distribution on each attribute at PETA dataset. Similar as RAP dataset, the \emph{example-based} evaluation could improve a lot using multi-attribute joint learning models than each single attribute alone. As we can see, the results on PETA dataset is usually higher than RAP dataset on both \emph{label-based} and \emph{example-based} evaluations.
This is reasonable because the PETA dataset is initially collected for person re-identification tasks and annotated based on identification, which means the same person will have the same attributes regardless the influence of the occlusion, viewpoint etc. In addition, the dataset is randomly divided into the training set, validation set and test set.
The same person's image may occur in both training set and test set, which is little inappropriate for attribute recognition tasks.

To take further research on the parts' influence on attribute recognition, we make an extra experiment that utilizing the fullbody and parts together to recognize attribute, which is named as DeepMAR* in Table~\ref{tab:accuracy_variant_evaluation}. We adopt the DeepMAR as the basic structure.
The input of DeepMAR has four blocks, including body image, head-shoulder image, upbody image and lowerbody image.
Then features are concatenated together in the FC7 layer.
After that, the weighed sigmoid cross entropy attribute loss \cite{li2015attribute} is adopted to train the whole network. The ground truth part annotations is used in test stage for analyze the influence of parts.
In the Table~\ref{tab:accuracy_variant_evaluation}, after fusing the body features and parts features together, both the \emph{label-based} and \emph{example-based} evaluation could further improve about 1 point.
This is another evidence that fusing parts and fullbody information could assist attribute recognition task.

%To verify the baseline methods' efficiency, we take further experiments on PETA dataset, which has been shown on Table \ref{tab:accuracy_variant_evaluation}.
%The features used in PETA for SVM model are the same to RAP. The experiment setting is the same as the paper \cite{deng2014pedestrian}.
%As we can see from Table \ref{tab:accuracy_variant_evaluation}, compared with multi-attribute jointly learning model DeepMAR, the SVM model achieves lower performance on both datasets.
%Especially, the multi-attribute jointly learning model could improve more than 25 percents than the SVM model using \emph{example-based} evaluation, including Accuracy and Precision, on both datasets.
%It shows better ability in the predictions' consistence on a pedestrian image.
%Generally the results on PETA dataset is usually higher than RAP dataset.
%This is reasonable because the PETA dataset is initially collected for person re-identification tasks and annotated based on identification, which means the same person will have the same attributes regardless the influence of the occlusion, viewpoint etc.
%In addition, the dataset is randomly divided into the training set, val set and test set.
%The same person's image may occur in both training set and test set, which is little inappropriate for attribute recognition tasks.

\begin{table}[!tbp]
%\vspace{-1em}
  \centering
  %\scriptsize
  \caption{Attribute recognition results(\%) on \textbf{RAP} and \textbf{PETA} dataset. The bold number is the
max value on each row except the last row.}
    \begin{tabular}{c|c|c|c|c|c|c|c|c|c|c}
    \hline
    \multirow{2}[4]{*}{Methods} & \multicolumn{5}{c|}{RAP}  & \multicolumn{5}{c}{PETA \cite{deng2014pedestrian}} \\
    \cline{2-11}
          & mA    & Accuracy & Precision & Recall & F1    & mA    & Accuracy & Precision & Recall & F1 \\
    \hline
    ELF-mm & 69.94  & 29.29  & 32.84  & 71.18  & 44.95  & 75.21  & 43.68  & 49.45  & 74.24  & 59.36  \\
    FC7-mm & 72.28  & 31.72  & 35.75  & 71.78  & 47.73  & 76.65  & 45.41  & 51.33  & 75.14  & 61.00  \\
    FC6-mm & 73.32  & 33.37  & 37.57  & 73.23  & 49.66  & 77.96  & 48.13  & 54.06  & 76.49  & 63.35  \\
    ACN \cite{sudowe2015person}   & 69.66  & \textbf{62.61} & \textbf{80.12}  & 72.26  & \textbf{75.98}  & 81.15  & 73.66  & \textbf{84.06}  & 81.26  & 82.64  \\
    DeepMAR \cite{li2015attribute} & \textbf{73.79} & 62.02 & 74.92  & \textbf{76.21}  & 75.56  & \textbf{82.89}  & \textbf{75.07}  & 83.68  & \textbf{83.14}  & \textbf{83.41}  \\
    DeepMAR* & 74.44  & 63.67  & 76.53  & 77.47  & 77.00  & -     & -     & -     & -     & - \\
    \hline
    \end{tabular}%
\label{tab:accuracy_variant_evaluation}
\vspace{-2em}
\end{table}%

%\begin{figure}[!tbp]
%\begin{center}
%\includegraphics[width=0.7\linewidth]{attachments/multi-attribute-results.pdf}
%\vspace{-1.8mm}
%\caption{Recognition results of the woman in the left. Both the DeepMAR and SVM has the same recall. However the SVM achieve higher }
%\label{tab:result_analysis}
%\end{center}
%\vspace{-3em}
%\end{figure}
% An intuitive concept about the recognition results. The red and green box is the groundtruth positive and negative label sets. The yellow and purple box is the predicted positive box using SVM and ACN/DeepMAR. Using the multi-attribute joint learning model could improve the negative recognition rate a lot.

\section{Conclusions}
This paper aims to improve the research of human attribute recognition in real surveillance scenarios.
To promote the research, we collect a large-scale richly annotated pedestrian (RAP) dataset from real scenarios.
The collected RAP dataset has 41,585 pedestrian images with different types of annotations, including viewpoints, occlusions, body parts and other common attributes, which is very useful for developing large-scale attribute recognition algorithms.
The benchmark results based on linear SVM model and multi-attribute joint learning model ACN and DeepMAR are shown for further comparison.
Finally, we make a quantitative analysis about the viewpoints, occlusions, body parts and interrelationship in attribute recognition and show that the environmental and contextual factors guided attribute recognition can obtain impressive improvements, which could be used to guide to recognize attributes.

\bibliographystyle{splncs03}
\bibliography{egbib}

\begin{thebibliography}{10}
\providecommand{\url}[1]{\texttt{#1}}
\providecommand{\urlprefix}{URL }

\bibitem{bourdev2011describing}
Bourdev, L., Maji, S., Malik, J.: Describing people: A poselet-based approach
  to attribute classification. In: Proc. ICCV (2011)

\bibitem{deng2014pedestrian}
Deng, Y., Luo, P., Loy, C.C., Tang, X.: Pedestrian attribute recognition at far
  distance. In: Proc. MM (2014)

\bibitem{donahue2013decaf}
Donahue, J., Jia, Y., Vinyals, O., Hoffman, J., Zhang, N., Tzeng, E., Darrell,
  T.: Decaf: A deep convolutional activation feature for generic visual
  recognition. arXiv preprint arXiv:1310.1531  (2013)

\bibitem{felzenszwalb2010object}
Felzenszwalb, P.F., Girshick, R.B., McAllester, D., Ramanan, D.: Object
  detection with discriminatively trained part-based models. TPAMI  32(9),
  1627--1645 (2010)

\bibitem{feris2014attribute}
Feris, R., Bobbitt, R., Brown, L., Pankanti, S.: Attribute-based people search:
  Lessons learnt from a practical surveillance system. In: Proc. ICMR (2014)

\bibitem{fogel1989gabor}
Fogel, I., Sagi, D.: Gabor filters as texture discriminator. Biological
  cybernetics  61(2),  103--113 (1989)

\bibitem{girshick2014rich}
Girshick, R., Donahue, J., Darrell, T., Malik, J.: Rich feature hierarchies for
  accurate object detection and semantic segmentation. In: Proc. CVPR (2014)

\bibitem{Gray2008reidentification}
Gray, D., Tao, H.: Viewpoint invariant pedestrian recognition with an ensemble
  of localized features. In: Proc. ECCV (2008)

\bibitem{hirzer2011person}
Hirzer, M., Beleznai, C., Roth, P.M., Bischof, H.: Person re-identification by
  descriptive and discriminative classification. In: Image Analysis, pp.
  91--102. Springer (2011)

\bibitem{jia2014caffe}
Jia, Y., Shelhamer, E., Donahue, J., Karayev, S., Long, J., Girshick, R.,
  Guadarrama, S., Darrell, T.: Caffe: Convolutional architecture for fast
  feature embedding. In: Proc. MM (2014)

\bibitem{krizhevsky2012imagenet}
Krizhevsky, A., Sutskever, I., Hinton, G.E.: Imagenet classification with deep
  convolutional neural networks. In: Proc. NIPS (2012)

\bibitem{kumar2011describable}
Kumar, N., Berg, A.C., Belhumeur, P.N., Nayar, S.K.: Describable visual
  attributes for face verification and image search. TPAMI  33(10) (2011)

\bibitem{layne2012towards}
Layne, R., Hospedales, T.M., Gong, S.: Towards person identification and
  re-identification with attributes. In: Proc. ECCV Workshops (2012)

\bibitem{layne2012person}
Layne, R., Hospedales, T.M., Gong, S., Mary, Q.: Person re-identification by
  attributes. In: Proc. BMVC (2012)

\bibitem{li2014clothing}
Li, A., Liu, L., Wang, K., Liu, S., Yan, S.: Clothing attributes assisted
  person re-identification. TCSVT  25(5),  869--878 (2014)

\bibitem{li2015attribute}
Li, D., Chen, X., Huang, K.: Multi-attribute learning for pedestrian attribute
  recognition in surveillance scenarios. In: Proc. ACPR (2015)

\bibitem{liu2012person}
Liu, C., Gong, S., Loy, C.C., Lin, X.: Person re-identification: What features
  are important? In: Proc. ECCV Workshops (2012)

\bibitem{liu2014deep}
Liu, Z., Luo, P., Wang, X., Tang, X.: Deep learning face attributes in the
  wild. arXiv preprint arXiv:1411.7766  (2014)

\bibitem{luo2013deep}
Luo, P., Wang, X., Tang, X.: A deep sum-product architecture for robust facial
  attributes analysis. In: Proc. ICCV (2013)

\bibitem{maji2008classification}
Maji, S., Berg, A.C., Malik, J.: Classification using intersection kernel
  support vector machines is efficient. In: Proc. CVPR (2008)

\bibitem{nortcliffe2011people}
Nortcliffe, T.: People analysis cctv investigator handbook. Home Office Centre
  of Applied Science and Technology  2, ~3 (2011)

\bibitem{ouyang2015deepattributes}
Ouyang, W., Li, H., Zeng, X., Wang, X.: Learning deep representation with
  large-scale attributes. In: Proc. ICCV (2015)

\bibitem{prosser2010person}
Prosser, B., Zheng, W.S., Gong, S., Xiang, T., Mary, Q.: Person
  re-identification by support vector ranking. In: Proc. BMVC (2010)

\bibitem{razavian2014cnn}
Razavian, A.S., Azizpour, H., Sullivan, J., Carlsson, S.: Cnn features
  off-the-shelf: an astounding baseline for recognition. In: Proc. CVPR
  Workshops (2014)

\bibitem{schmid2001constructing}
Schmid, C.: Constructing models for content-based image retrieval. In: Proc.
  CVPR (2001)

\bibitem{scholkopf2002learning}
Sch{\"o}lkopf, B., Smola, A.J.: Learning with kernels: Support vector machines,
  regularization, optimization, and beyond. MIT press (2002)

\bibitem{shao2015deeply}
Shao, J., Kang, K., Loy, C.C., Wang, X.: Deeply learned attributes for crowded
  scene understanding. In: Proc. CVPR. pp. 4657--4666 (2015)

\bibitem{sudowe2015person}
Sudowe, P., Spitzer, H., Leibe, B.: Person attribute recognition with a
  jointly-trained holistic cnn model. In: Proc. ICCV Workshops. pp. 87--95
  (2015)

\bibitem{taigman2014deepface}
Taigman, Y., Yang, M., Ranzato, M., Wolf, L.: Deepface: Closing the gap to
  human-level performance in face verification. In: Proc. CVPR (2014)

\bibitem{tian2014pedestrian}
Tian, Y., Luo, P., Wang, X., Tang, X.: Pedestrian detection aided by deep
  learning semantic tasks. In: Proc. CVPR (2015)

\bibitem{vaquero2009attribute}
Vaquero, D., Feris, R.S., Tran, D., Brown, L., Hampapur, A., Turk, M., et~al.:
  Attribute-based people search in surveillance environments. In: Proc. WACV
  Workshops (2009)

\bibitem{zhang2014review}
Zhang, M.L., Zhou, Z.H.: A review on multi-label learning algorithms. Knowledge
  and Data Engineering, IEEE Transactions on  26(8),  1819--1837 (2014)

\bibitem{zhang2014panda}
Zhang, N., Paluri, M., Ranzato, M., Darrell, T., Bourdev, L.: Panda: Pose
  aligned networks for deep attribute modeling. In: Proc. CVPR (2014)

\bibitem{zhu2013pedestrian}
Zhu, J., Liao, S., Lei, Z., Yi, D., Li, S.Z.: Pedestrian attribute
  classification in surveillance: Database and evaluation. In: Proc. ICCV
  Workshops (2013)

\bibitem{zhu2015multi}
Zhu, J., Liao, S., Yi, D., Lei, Z., Li, S.Z.: Multi-label cnn based pedestrian
  attribute learning for soft biometrics. In: Proc. ICB (2015)

\end{thebibliography}
\end{document}